\documentclass[11pt]{article}

\usepackage[a4paper,margin=1in]{geometry}
\usepackage[utf8]{inputenc}
\usepackage[T1]{fontenc}
\usepackage[protrusion=true,expansion=false]{microtype}
\usepackage{amsmath,amssymb,amsfonts}
\usepackage{booktabs}
\usepackage{tabularx}
\usepackage{longtable}
\usepackage{array}
\usepackage{multirow}
\usepackage{makecell}
\usepackage{graphicx}
\usepackage{xcolor}
\usepackage[hidelinks,colorlinks=true,linkcolor=blue!50!black,citecolor=blue!50!black,urlcolor=blue!50!black]{hyperref}
\usepackage{url}
\usepackage{enumitem}
\usepackage{listings}
\usepackage{caption}
\usepackage{titlesec}
\usepackage{natbib}
\usepackage{authblk}
\usepackage{float}
\usepackage{flafter}
\usepackage[section]{placeins}
\usepackage{subcaption}
\graphicspath{{figures/}}

\titlespacing*{\section}{0pt}{1.5ex plus .2ex}{0.8ex plus .2ex}
\titlespacing*{\subsection}{0pt}{1.2ex plus .2ex}{0.6ex plus .2ex}
\setlist{nosep}
\newcommand{\router}{\textsc{SCX Router}}

\newcommand{\best}{\textbf}

\title{\textbf{SCX Router: Streaming Zero-Shot Model Selection with a Decoder-KV Classifier and a Real-World Task Ontology}}
\author[1]{Ihor Stepanov}
\author[2]{Aleksandr Smechov}
\author[1]{Mykhailo Shtopko}
\author[1]{Dmytro Vodianytskyi}
\author[1]{Oleksandr Lukashov}
\affil[1]{Knowledgator}
\affil[2]{SCX.ai Holdings Limited}
\date{}

\begin{document}
\maketitle

\begin{abstract}
\noindent
The rapid proliferation of large language models (LLMs) and the growing diversity of their applications presents a unique optimization opportunity: selecting the right model for the task, while optimizing for speed, cost, and quality at a per-task level. However, inference endpoints can vary widely in quality, price, latency, context support, tool use, domain expertise, and reasoning behavior. This heterogeneity makes manual heuristics difficult to maintain and unlikely to achieve consistently favorable speed--cost--quality trade-offs on their own. We introduce \router{}, a lightweight GLiClass-based router that assigns a suitability score to each inference-time model label without autoregressive generation. The released 0.6B-parameter checkpoint\footnote{Checkpoint: \url{https://huggingface.co/scx-admin/scx-router-v0.1}. Code: \url{https://github.com/Knowledgator/GLiClass-Model-Router}.} (Apache 2.0) combines a Qwen3 decoder with a shallow bidirectional scorer. Its decoder-KV execution path preserves a text-only key--value cache across a session, encodes only new dialogue turns, and evaluates transient candidate-label tokens without adding them to the persistent cache. The same checkpoint also predicts task type, difficulty, reasoning mode, and expected output length, and supports custom zero-shot labels. For task generation, we construct a task ontology with 23 families, 115 task types, 345 routable subtypes, 1,173 synthetic examples, and an orthogonal axis of 30 domains. Using this structure, we generate 150,000 verifier-scored tasks and 15,000 open-ended tasks. We then train the Qwen3 decoder on these tasks, while explicitly separating learned request prediction from per-task policies for attributes such as eligibility, cost, cache reuse, safety, and sovereignty. Across six LiveBench subsets, the router outperforms the mean candidate; on the selected 1,000-task subset, it achieves an aggregate top-1 score of 0.707 versus 0.696 for the strongest fixed model, with benchmark-dependent gains.
\end{abstract}


% ========================
%        SECTION 1
% ========================
\section{Introduction}
\label{sec:intro}

Closed-source, frontier LLMs continue to fight for dominance on general intelligence leaderboards\footnote{\url{https://arcprize.org/leaderboard}}. But open-weights model families vary widely in their ability to reason, code, follow instructions, cover multiple languages, and more, while also ranging in price, latency, context length, tool access, and operational constraints. This gives users the advantage of choice from a large pool of intelligence, but also presents a dillema: how do you choose the right model for the task? Prior routing systems learn a binary strong-versus-weak decision, estimate answer quality, or cascade from inexpensive to expensive models~\citep{chen2023frugalgpt,aggarwal2023automix,ong2024routellm,dekoninck2024unified}. RouterBench formalizes the broader multi-model setting and demonstrates the value---and difficulty---of evaluation on a shared outcome matrix~\citep{hu2024routerbench}. In production, the problem is harder: candidate rosters change, dialogue histories grow, cache reuse changes the incremental price of switching, and some requests require hard policy constraints that cannot be reduced to a scalar preference.

\router{} is designed for such settings. It adapts GLiClass~\citep{stepanov2025gliclass}, a generalist label--text classifier derived from GLiNER~\citep{zaratiana2023gliner}, to a causal decoder with a persistent KV cache. Instead of prompting another LLM to emit a route token, the router jointly processes a request and natural-language candidate labels and returns one logit per label in a non-generative forward pass. Because labels are inputs rather than fixed output neurons, the same interface supports model names, task taxonomies, difficulty levels, reasoning modes, output-length buckets, safety signals, and deployment-specific labels.

The architectural choice matters most in dialogue. A stateless encoder must process the full conversation whenever a new message arrives. A generative router may reuse its prefix cache but must still decode and parse route tokens. The decoder-KV path in \router{} keeps only request-context keys and values between turns. New request tokens extend that cache; candidate labels are then scored against the cached context and their KV tensors are discarded. This yields a streaming classifier rather than a second conversational generator.

Model scores alone are not a routing policy. We therefore keep the learned predictor separate from a deterministic decision layer. The latter can exclude endpoints that violate context, tool, modality, residency, privacy, or safety requirements, then trade predicted performance against incremental token cost, average latency, and cache state. This separation also makes empirical model-performance profiles replaceable without retraining the semantic classifier.

\subsection{Contributions}
\label{sec:contributions}

We make the following contributions.

\begin{enumerate}[leftmargin=*,itemsep=3pt]
  \item \textbf{A streaming zero-shot router.} We present a decoder-KV GLiClass architecture that combines inference-time label semantics, a chat-aligned causal backbone, a non-generative classification head, and text-only session-cache reuse.
  \item \textbf{A multi-signal routing interface.} One checkpoint estimates model suitability, task type, difficulty, reasoning mode, expected output length, and arbitrary custom labels. Learned prediction remains separate from hard eligibility and business policy.
  \item \textbf{A task ontology and synthetic data suite.} We construct an ontology containing 23 task families, 115 task types, 345 routable subtypes, 1,173 synthetic examples, and an orthogonal axis of 30 domains. It structures 150,000 verifier-scored and 15,000 \texttt{gpt5.6-sol}-judged synthetic tasks with real-world workflow structure.
  \item \textbf{A taxonomy of model-routing patterns.} We introduce direct endpoint routing, attribute-mediated performance routing, hybrid constrained routing, and hierarchical planner--worker routing for agentic workflows, while distinguishing released paths, implemented components, and proposed compositions.
  \item \textbf{An evaluation of routing quality.} We evaluate the released classification heads and distinguish mean-candidate from fixed-model end-to-end baselines. For the expanded 11-endpoint collection, observation masks preserve unequal coverage and prevent missing outcomes from becoming negative labels.
\end{enumerate}

The released checkpoint is available on Hugging Face, and the implementation and benchmark harness are public. (Checkpoint: \url{https://huggingface.co/scx-admin/scx-router-v0.1}. Code: \url{https://github.com/Knowledgator/GLiClass}.)

% ========================
%        SECTION 2
% ========================
\section{Related Work}
\label{sec:related}

\subsection{Routing and Cascading Across LLMs}

FrugalGPT learns cascades under budget constraints and shows that heterogeneous API models can improve the cost--quality frontier~\citep{chen2023frugalgpt}. AutoMix similarly escalates based on self-verification and a partially observable decision process~\citep{aggarwal2023automix}. RouteLLM learns strong-versus-weak routing from preference data and studies transfer when the underlying pair changes~\citep{ong2024routellm}. More recent work unifies routing and cascading and identifies quality estimation as the central statistical problem~\citep{dekoninck2024unified}. RouterBench supplies more than 405,000 inference outcomes and emphasizes common task coverage for trustworthy comparisons~\citep{hu2024routerbench}.

\router{} differs along three axes. First, it performs multi-label suitability prediction over a configurable roster rather than a fixed binary gate. Second, natural-language labels allow semantic transfer to new candidates and auxiliary taxonomies. Third, the persistent state belongs to a classifier built on a causal decoder: it can reuse dialogue tokens without generating a route. Semantic zero-shot transfer does not eliminate the need for outcome data, however; a new endpoint name may be accepted mechanically while remaining poorly calibrated empirically.

\subsection{Efficient Zero-Shot Classification}

GLiNER jointly represents text and entity labels for generalist named-entity recognition~\citep{zaratiana2023gliner}. GLiClass extends this paradigm to sequence classification, supporting zero- and few-shot classification while avoiding one text--label forward pass per class~\citep{stepanov2025gliclass}. The original GLiClass family primarily uses bidirectional encoders. \router{} retains its label-conditioned scorer but places it after a Qwen3 causal backbone~\citep{qwen3technicalreport}. A shallow DeBERTa-style encoder~\citep{he2020deberta} restores bidirectional interaction over the transient label suffix before classification.

Embedding and bi-encoder methods precompute cheap label representations but expose limited token-level label--text interaction. Pairwise cross-encoders offer stronger interaction at a cost proportional to the number of labels. Prompted generative routers are flexible and chat-native, but add decoding latency and output-format variance. The decoder-KV design occupies a middle point: labels remain dynamic, the backbone is causal and cacheable, and the output is discriminative.

\subsection{Real-World and Agentic Evaluation}

Static question answering is an incomplete proxy for routing real applications. GAIA emphasizes questions that combine reasoning, browsing, multimodality, and tools~\citep{mialon2023gaia}; AgentBench evaluates LLMs across interactive environments~\citep{liu2023agentbench}; SWE-bench requires repository-scale code modification and test execution~\citep{jimenez2023swebench}; and OSWorld pairs realistic tasks with initial-state configurations, files, and execution-based evaluators~\citep{xie2024osworld}. These benchmarks motivate our task packages: the prompt alone is insufficient for many workloads, so generation includes the context, artifacts, criteria, and checks needed to evaluate an outcome.

Open-ended evaluation often relies on LLM judges. MT-Bench documents both their scalability and systematic position, verbosity, and self-enhancement biases~\citep{zheng2023judging}. Panels of heterogeneous judges can reduce single-model bias~\citep{verga2024replacing}. We therefore treat deterministic verifiers as preferable when semantics permit and require judge identity, prompt, rubric, and coverage metadata for subjective tasks.

% ========================
%        SECTION 3
% ========================
\section{Task Formulation}
\label{sec:formulation}

Let $x_t$ be the new request content at turn $t$, $h_{<t}$ the preceding conversation, and $\mathcal{M}_t=\{m_1,\ldots,m_K\}$ the available candidate-label vocabulary before product constraints. The classifier computes a logit $a_{t,k}$ and probability-like score
\begin{equation}
  p_{t,k}=\sigma(a_{t,k})
\end{equation}
for each candidate in the multi-label view. A candidate is emitted when $p_{t,k}\geq\tau$, with a default threshold of $0.5$ in the released pipeline. For a single-label auxiliary task, softmax normalization and an argmax decision are used instead.

Routing supervision is derived from observed downstream outcomes. For task $i$ and model $m$, let $s_{i,m}\in[0,1]$ denote a benchmark-normalized score and $o_{i,m}\in\{0,1\}$ denote whether the outcome was actually observed. Our dataset builder defines the positive set as models tied for the highest observed score, optionally within a tolerance $\epsilon$:
\begin{equation}
  Y_i=\left\{m\in\mathcal{M}_i:
  s_{i,m}\geq \max_{j\in\mathcal{M}_i}s_{i,j}-\epsilon\right\}.
  \label{eq:winners}
\end{equation}
Equation~\ref{eq:winners} operationalizes success from the best observed outcomes and avoids imposing one universal threshold across heterogeneous benchmark evaluators.

If every candidate is evaluated on every task, $Y_i$ supports a paired routing objective. When coverage differs, $o_{i,m}=0$ is missing data, not a failed response. A valid loss must mask unobserved pairs,
\begin{equation}
 \mathcal{L}_{\mathrm{masked}}=
 -\frac{1}{\sum_{i,m}o_{i,m}}
 \sum_{i,m}o_{i,m}\left[y_{i,m}\log p_{i,m}+
 (1-y_{i,m})\log(1-p_{i,m})\right],
 \label{eq:masked}
\end{equation}
or restrict training and evaluation to a paired intersection. The eleven-endpoint collection has unequal endpoint counts, so this paper does not infer unobserved failures or report 11-way comparative accuracy.

The final route is a policy decision. Let $\mathcal{E}_t\subseteq\mathcal{M}_t$ be the candidates remaining after context, tool, modality, region, privacy, and safety constraints. A policy chooses
\begin{equation}
  m_t^*=\arg\max_{m\in\mathcal{E}_t} U_t(m),
\end{equation}
where $U_t$ can combine request-specific scores, empirical performance profiles, expected incremental cost, latency, and cache reuse. Section~\ref{sec:policy} gives the implemented cost-aware form.

% ========================
%        SECTION 4
% ========================
\section{Decoder-KV Router}
\label{sec:architecture}

The released checkpoint is an approximately 0.6B-parameter GLiClass model with a Qwen3-0.6B causal backbone~\citep{qwen3technicalreport}. The backbone has 28 decoder layers, hidden size 1,024, 16 query heads and 8 KV heads; training uses sequences up to 4,096 tokens. Three special tokens organize the classification sequence: \texttt{<<LABEL>>}, \texttt{<<SEP>>}, and \texttt{<<EXAMPLE>>}. The scorer is a two-layer DeBERTa-v2 encoder without its own embedding layer, followed by projections and a shared MLP of widths $2h\rightarrow1024\rightarrow512\rightarrow1$.

\begin{figure}[!htbp]
\centering
\includegraphics[width=\textwidth]{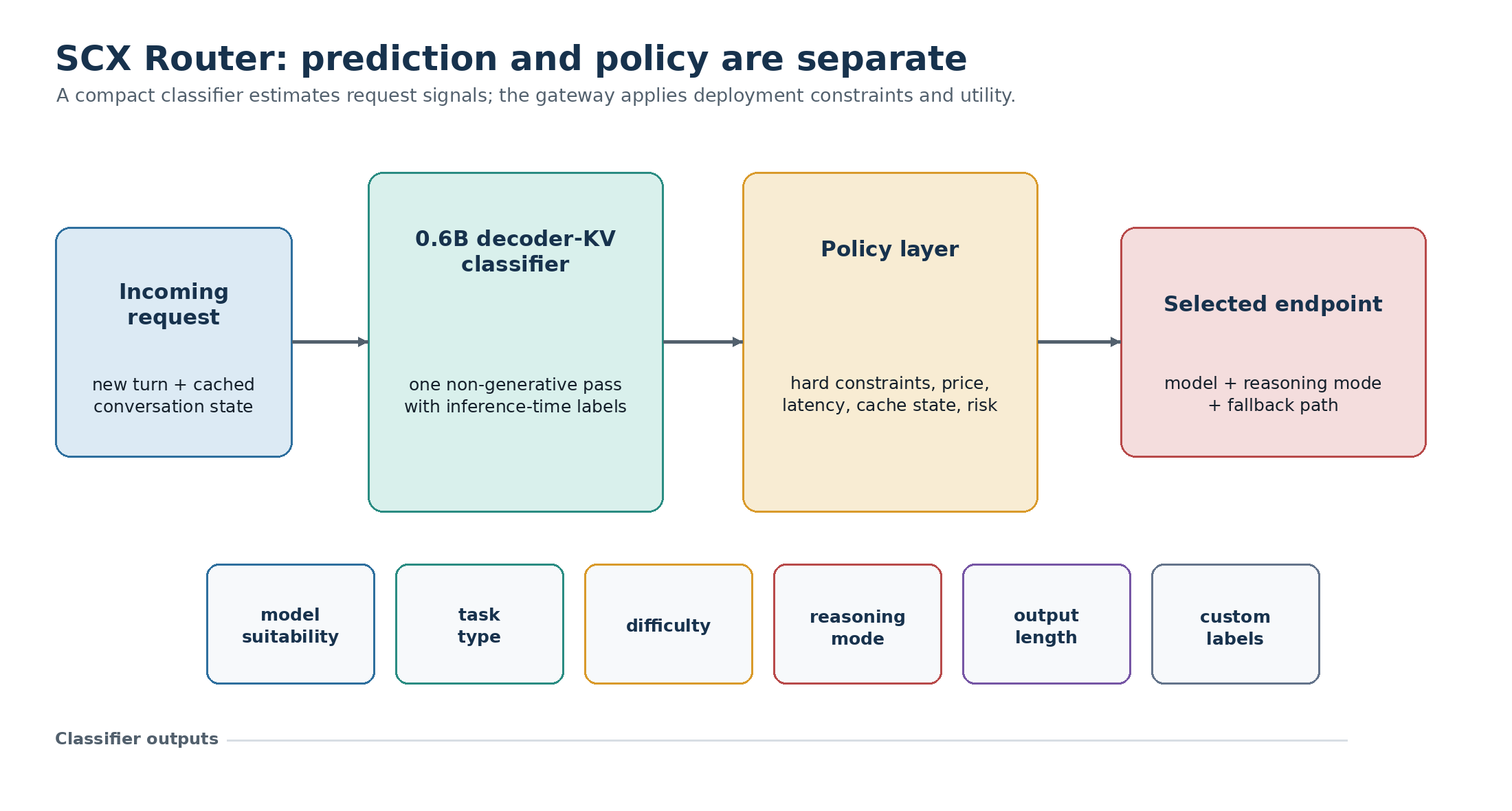}
\caption{System overview. The classifier returns request signals; a separate application policy applies hard eligibility constraints and selects an endpoint. The displayed signal families use the same label-conditioned inference interface.}
\label{fig:system}
\end{figure}

Figure~\ref{fig:system} summarizes the separation between learned request signals and the deployment policy that selects an eligible endpoint.

\subsection{Sequence and Scoring}

The canonical implementation formats the persistent context and transient labels as
\begin{align}
 c_{\leq t} &= \operatorname{format}(\text{prompt},\text{examples},h_{<t},x_t),\\
 q(\mathcal{L}) &= \texttt{<<SEP>>}\;\ell_1\;\texttt{<<LABEL>>}\;\cdots\;
 \ell_K\;\texttt{<<LABEL>>}\;\texttt{<<SEP>>}.
 \label{eq:format}
\end{align}
The label text therefore precedes its \texttt{<<LABEL>>} marker, matching the data-processing and streaming implementation.

For a stateless call, the causal decoder processes $[c_{\leq t};q(\mathcal{L})]$. For a session, let $\Delta c_t$ denote only the persistent-context tokens appended since the last cached call. Those new tokens update a persistent text-only cache,
\begin{equation}
 (H_t^{x},K_t,V_t)=D_\theta(\Delta c_t;K_{t-1},V_{t-1}),
\end{equation}
and label tokens are evaluated conditionally,
\begin{equation}
 H_t^{\ell}=D_\theta(q(\mathcal{L});K_t,V_t).
\end{equation}
The label-stage keys and values are not written back to $(K_t,V_t)$. Thus repeated classification can change the label roster without contaminating dialogue state.

The transient hidden states pass through a bidirectional scorer encoder $B_\phi$:
\begin{equation}
 Z=B_\phi(H_t^{\ell}).
\end{equation}
The scorer extracts a joint representation $z_t$ from the final separator and label representations $z_k$ from the \texttt{<<LABEL>>} positions. Because $B_\phi$ is bidirectional, the final separator summarizes the full transient label section as well as decoder-conditioned request context; it is not simply the last request-token state. Projected request and label representations are concatenated and scored:
\begin{align}
 \tilde z_t &= W_t z_t, & \tilde z_k &= W_\ell z_k,\\
 a_{t,k} &= \operatorname{MLP}_\psi([\tilde z_t;\tilde z_k]).
\end{align}

Two properties follow from this persistent--transient split. First, cache reuse amortizes request-side representation, not the entire routing decision. If a turn contributes $\Delta n$ context tokens and the current roster occupies $L$ label tokens, the cached path extends the state with $\Delta n$ tokens and, when triggered, executes the transient $L$-token suffix and scorer. The suffix still attends to retained history, so latency can depend on both cache length and roster size; reuse avoids recomputing earlier hidden states but does not make routing cost independent of history. Second, labels are in-band inputs rather than fixed output neurons: the same projections and MLP score every marker. The head therefore admits a variable roster and independent suitability scores for multiple acceptable endpoints. Label shuffling during training discourages position-specific shortcuts, but a new label is only semantically scorable---not automatically calibrated---until supported by outcome evidence.

\begin{figure}[H]
\centering
\includegraphics[width=\textwidth]{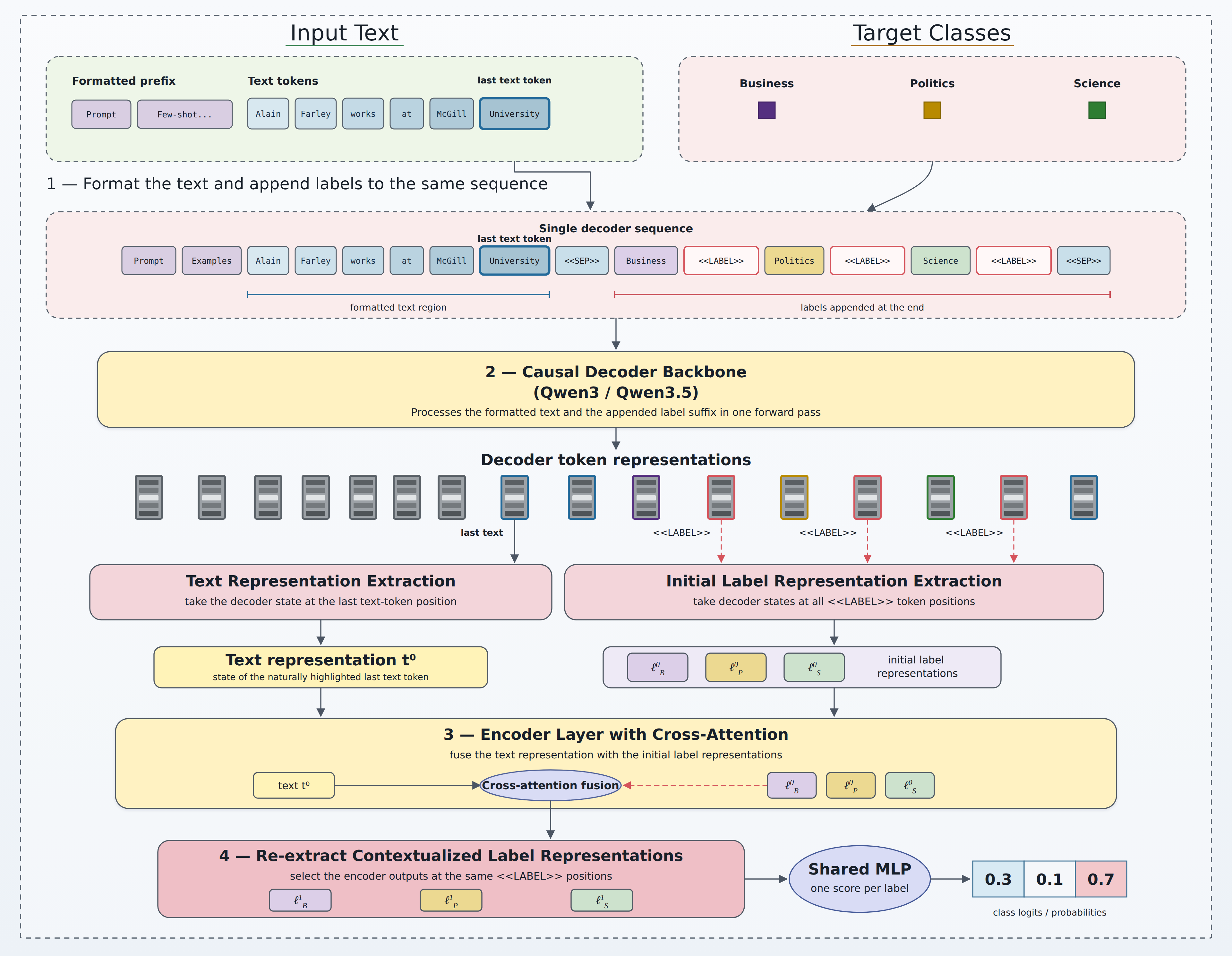}
\caption{GLiClass decoder architecture. A causal decoder processes the text and appended class labels in one sequence. The final text-token state and label-marker states are extracted, fused through cross-attention, and scored by a shared per-label MLP.}
\label{fig:architecture}
\end{figure}

Figure~\ref{fig:architecture} visualizes the complete text--label scoring path.

\FloatBarrier
\subsection{Streaming Execution}

Each session owns a cache and a classification trigger. The implementation supports classification on every chunk, after $n$ tokens, on a delimiter, never, or over a sliding window. Sessions of unequal cache length are left-padded for batched decoder calls, with masks and explicit positions preserving logical token locations. When physical cache truncation is enabled, absolute rotary positions continue increasing rather than reusing cropped positions. Optional CPU offload bounds GPU cache memory. At the service layer, TTL, LRU capacity, maximum cache length, explicit reset, and deletion govern cache lifetime.

Cache update and classification are deliberately decoupled. A nonempty chunk can advance persistent state even when the trigger suppresses scoring; the next triggered call then sees that accumulated context. Trigger timing and context selection are also separate: a periodic or delimiter trigger can be composed with a bounded recent window. This makes the deployment trade-off explicit. End-of-turn triggers reduce redundant intermediate decisions, while per-chunk triggers provide earlier route changes; windows and physical truncation bound active context at the cost of discarding older evidence.

The router cache is distinct from the prompt cache of the selected generative endpoint. It amortizes repeated classification only; after a route change, the destination may still need to replay the conversation. We therefore expose downstream cache state to the policy instead of treating router-side reuse as evidence that switching is free.

Table~\ref{tab:architecture_comparison} contrasts this design with common alternatives.

\begin{table}[!htbp]
\centering
\small
\caption{Qualitative comparison of router architectures. ``Dynamic labels'' means that candidate label text can change at inference time.}
\label{tab:architecture_comparison}
\begin{tabularx}{\textwidth}{lXXXX}
\toprule
\textbf{Approach} & \textbf{Dynamic labels} & \textbf{New dialogue turn} & \textbf{Generation} & \textbf{Primary trade-off} \\
\midrule
Embedding / bi-encoder & native & usually re-embed full context & no & cheapest interaction, weaker cross-text reasoning \\
Encoder classifier / NLI & fixed head or pairwise & re-encode full context & no & strong discrimination, no token-level session cache \\
Generative LLM router & prompt-native & prefix cache may be reused & yes & flexible and explainable, but decoding and parsing required \\
\router{} decoder-KV & native & encode only new text; re-score labels & no & compact discriminative path with persistent dialogue state \\
\bottomrule
\end{tabularx}
\end{table}

% ========================
%        SECTION 5
% ========================
\section{Task Ontology}
\label{sec:ontology}

\begin{figure}[!htbp]
\centering
\includegraphics[width=\textwidth]{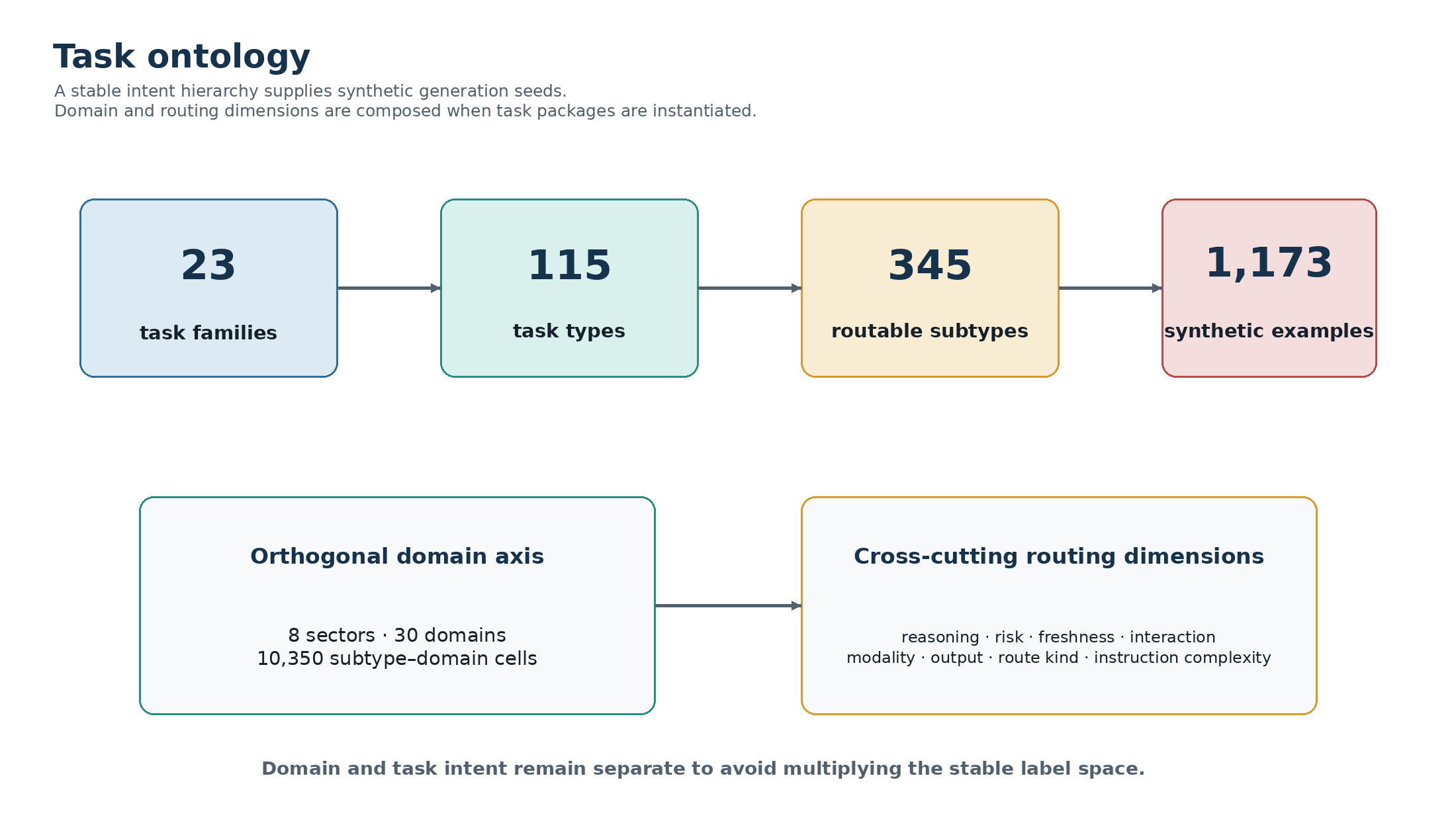}
\caption{Ontology structure. Task intent is modeled independently from domain and routing dimensions. Composing 345 subtypes with 30 domains defines 10,350 initial design cells, but domain-specific examples are instantiated only when needed.}
\label{fig:ontology}
\end{figure}

We construct the task ontology as a three-level hierarchy of intent: family, task type, and routable subtype. It contains 23 families, 115 task types, and 345 subtypes. Every family contains five task types and 15 subtypes. We additionally generate 1,173 synthetic example specifications to define coverage and seed downstream task generation; these examples are not benchmark prompts. An orthogonal domain axis and cross-cutting routing dimensions allow the same intent hierarchy to describe different application settings without multiplying the stable task labels. Figure~\ref{fig:ontology} depicts these axes and their composition.

\subsection{Intent, Domain, and Cross-Cutting Dimensions}

We separate semantic intent from context to avoid an unstable Cartesian label vocabulary. ``Summarize a clinical handoff'' and ``summarize a deployment incident'' share an operation but differ in domain, risk, artifacts, and likely endpoint requirements. The domain axis contains eight sectors and 30 domains. We bind a subtype to a domain scenario only when constructing an example or downstream record.

We define eight cross-cutting dimensions: reasoning level (1--5), risk (low through critical), freshness, interaction mode, input modality, output mode, route kind, and instruction complexity. These dimensions need not be mutually exclusive task classes. In particular, multi-turn and instruction-following behavior can apply to any semantic family, so we represent them as interaction and constraint dimensions rather than forcing them into the top-level intent axis.

We also define boundaries between historically overlapping labels. Analysis interprets evidence, reasoning derives conclusions, and problem solving proposes and validates a remedy. Comparison enumerates differences, evaluation scores against criteria, critique identifies weaknesses and improvements, and verification checks truth or acceptance evidence. Retrieval returns sources, question answering returns an answer, and fact checking returns an evidence-backed claim status. These distinctions target the weakest classes in the released 28-way task-type evaluation, where conceptually overlapping labels exhibit the largest confusion.

\FloatBarrier
Table~\ref{tab:ontology_counts} summarizes the ontology dimensions and their roles.

\begin{table}[!htbp]
\centering
\small
\caption{Summary of the task ontology.}
\label{tab:ontology_counts}
\begin{tabular}{lrp{7.5cm}}
\toprule
\textbf{Axis} & \textbf{Count} & \textbf{Role} \\
\midrule
Task families & 23 & broad semantic operation, such as retrieval, reasoning, programming, or agentic execution \\
Task types & 115 & operational subdivisions; five per family \\
Routable subtypes & 345 & stable fine-grained intents; three per task type \\
Synthetic example specifications & 1,173 & generated coverage examples and task-generation seeds \\
Domain sectors & 8 & broad economic or social contexts \\
Domains & 30 & reusable contexts composed with task subtypes \\
Subtype--domain design cells & 10,350 & possible initial single-turn generation jobs \\
Cross-cutting dimensions & 8 & reasoning, risk, freshness, interaction, modality, output, route kind, and instruction complexity \\
\bottomrule
\end{tabular}
\end{table}

% ========================
%        SECTION 6
% ========================
\section{Data Sources and Supervision}
\label{sec:data}

We use two sources of routing supervision. The initial corpus is benchmark-derived: candidate models are evaluated on training subsets of existing benchmarks. The second corpus is generated synthetically from the task ontology. Here \emph{real-world} describes task structure---dialogue state, tools, files, repository context, and acceptance criteria---rather than production user-log provenance. Table~\ref{tab:data_inventory} separates prompt origin from label meaning.

\begin{table}[!htbp]
\centering
\small
\setlength{\tabcolsep}{4pt}
\caption{Data inventory by origin and supervision. Broad and focused counts are training records; synthetic counts are generated task instances.}
\label{tab:data_inventory}
\begin{tabularx}{\textwidth}{>{\raggedright\arraybackslash}p{0.18\textwidth}>{\raggedright\arraybackslash}p{0.23\textwidth}>{\raggedright\arraybackslash}p{0.16\textwidth}>{\raggedright\arraybackslash}X}
\toprule
\textbf{Dataset or signal} & \textbf{Origin} & \textbf{Scale} & \textbf{Supervision and meaning} \\
\midrule
Benchmark model routing & Existing benchmark train subsets evaluated with candidate models & $\sim$104.5k broad; 22,303 focused & Observed outcomes identify the best-performing model or tied models. \\
Task type & Benchmark prompts classified by LLMs & $\sim$105.3k broad; 15,015 focused & Semantic operation requested by the task, independent of model identity. \\
Difficulty & Candidate outcomes on the same tasks & $\sim$104.6k broad; 17,318 focused & Five-level, roster-relative solvability: tasks solved by fewer models are harder. \\
Synthetic, verifier-scored & Ontology-conditioned synthetic task packages & 150,000 tasks & Deterministic, executable, or environment-based checks score candidate results. \\
Synthetic, judge-scored & Ontology-conditioned synthetic open-ended tasks & 15,000 tasks & \texttt{gpt5.6-sol} scores outputs against explicit task criteria. \\
\bottomrule
\end{tabularx}
\end{table}

\subsection{Benchmark-Derived Routing Data}

The released routing data begins with prompts from the training partitions of existing benchmarks spanning classification, question answering, reasoning, code, summarization, translation, instruction following, multi-turn interaction. We execute the candidate models on each prompt and retain the score and observation indicator for every attempted endpoint. Equation~\ref{eq:winners} converts these observed outcomes into model-routing targets; endpoint names or descriptions do not determine the target.

Task type and difficulty have distinct semantics. LLM annotators classify each benchmark prompt into the task-type vocabulary according to the requested operation, rather than its source benchmark. Difficulty is assigned after model evaluation from the relative ability of the candidate roster to solve the task. It is therefore roster- and evaluator-dependent: tasks solved by many models are easier, whereas tasks solved by few models are harder. A change in candidate roster or success rule can change the difficulty label.

The released checkpoint uses a broad 524,035-record GLiClass mixture followed by a focused 65,099-record routing mixture. The focused stage contains 22,303 model-routing, 17,318 difficulty, 15,015 task-type, 1,797 expected-output-length, 890 reasoning-mode, and approximately 7,700 propositional-logic records. These are classification-record counts, not counts of unique benchmark prompts. General topic, hallucination, guardrail, sentiment, emotion, NLI, toxicity, safety, and QA-domain data in the broad mixture preserve zero-shot classification ability.

\subsection{Ontology-Driven Synthetic Data}

The task ontology supplies task family, task type, subtype, domain, and interaction constraints to the generator. Every task instantiated through this path is synthetic. Depending on the environment, a package contains a prompt, dialogue or execution context, files, tool interfaces, repository state, assessment criteria, and an evaluator. This covers single- and multi-turn text, tool use, coding, repository-level work, agentic workflows, and multi-agent tasks. The initial generated difficulty is metadata only; the training difficulty label is recomputed from relative model outcomes.

We generated 150,000 tasks whose results can be verified with deterministic, executable, or environment-based checks. We generated a further 15,000 open-ended tasks for which deterministic verification is unsuitable; \texttt{gpt5.6-sol} evaluates their candidate outputs against the generated criteria. The judge identity, criteria, task package, candidate response, and score are retained as evaluation provenance. Together these pools contain 165,000 synthetic tasks. Agentic packages follow the reproducibility principle of OSWorld and SWE-bench: success depends on initial state and executable or rubric-based evaluation, not on an isolated prompt~\citep{xie2024osworld,jimenez2023swebench}.

\subsection{Outcome Coverage}

The expanded evaluation uses eleven endpoint identifiers, but endpoints were evaluated on different numbers of tasks. The 165,000-task total therefore does not imply a complete $165{,}000\times11$ outcome matrix. Every task--model record retains the observation flag $o_{i,m}$, and only observed outcomes contribute to routing labels, difficulty estimates, or performance profiles. Missing evaluations are not failures. Raw model averages over different task mixtures are not directly comparable, so an 11-endpoint leaderboard requires a paired task intersection or an explicit missing-outcome estimator.

% ========================
%        SECTION 7
% ========================
\section{Training}
\label{sec:training}

The released decoder-KV model is trained for multi-label classification with shuffled label order. For each row, only labels in \texttt{all\_labels} are tokenized into the suffix, and binary targets indicate the selected winners. The implementation supports standard binary cross entropy when focal parameters are disabled, as well as configurable focal modulation~\citep{lin2017focal}. The exact focal arguments, optimizer state, and complete launch command are not available, so we leave those fields unspecified.

Table~\ref{tab:training} summarizes the available training configuration.

\begin{table}[!htbp]
\centering
\small
\caption{Training configuration for the released checkpoint. Unavailable fields are marked accordingly.}
\label{tab:training}
\begin{tabular}{ll}
\toprule
\textbf{Hyperparameter} & \textbf{Value} \\
\midrule
Backbone & Qwen3-0.6B \\
Architecture & decoder-KV GLiClass \\
Problem type & multi-label classification \\
Maximum sequence length & 4,096 \\
Scorer encoder layers / heads & 2 / 16 \\
Scorer MLP hidden size & 1,024 \\
Dropout / scorer attention dropout & 0.1 / 0.1 \\
Stage-1 epochs / steps & 3 / 18,300 \\
Stage-1 per-device batch size & 4 \\
Stage-1 schedule & linear with warmup \\
Stage-2 decoder / scorer learning rate & $10^{-6}$ / $10^{-6}$ \\
Precision & bfloat16 \\
Gradient checkpointing & enabled \\
Train/evaluation split & 90/10 \\
Seed / label shuffling & 42 / enabled \\
Training hardware, global batch, wall time & not available \\
\bottomrule
\end{tabular}
\end{table}

Broad pre-training reaches flattened weighted binary accuracy 0.9273, F1 0.9241, precision 0.9238, recall 0.9273, and loss 0.218. These are label-decision metrics rather than example-level exact-match routing accuracy; many negative model labels can dominate them. We therefore use the routing-family and downstream results in Section~\ref{sec:evaluation} as the more relevant evidence.

% ========================
%        SECTION 8
% ========================
\section{Model-Routing Patterns}
\label{sec:routing_patterns}

\begin{figure}[!htbp]
\centering
\includegraphics[width=\textwidth]{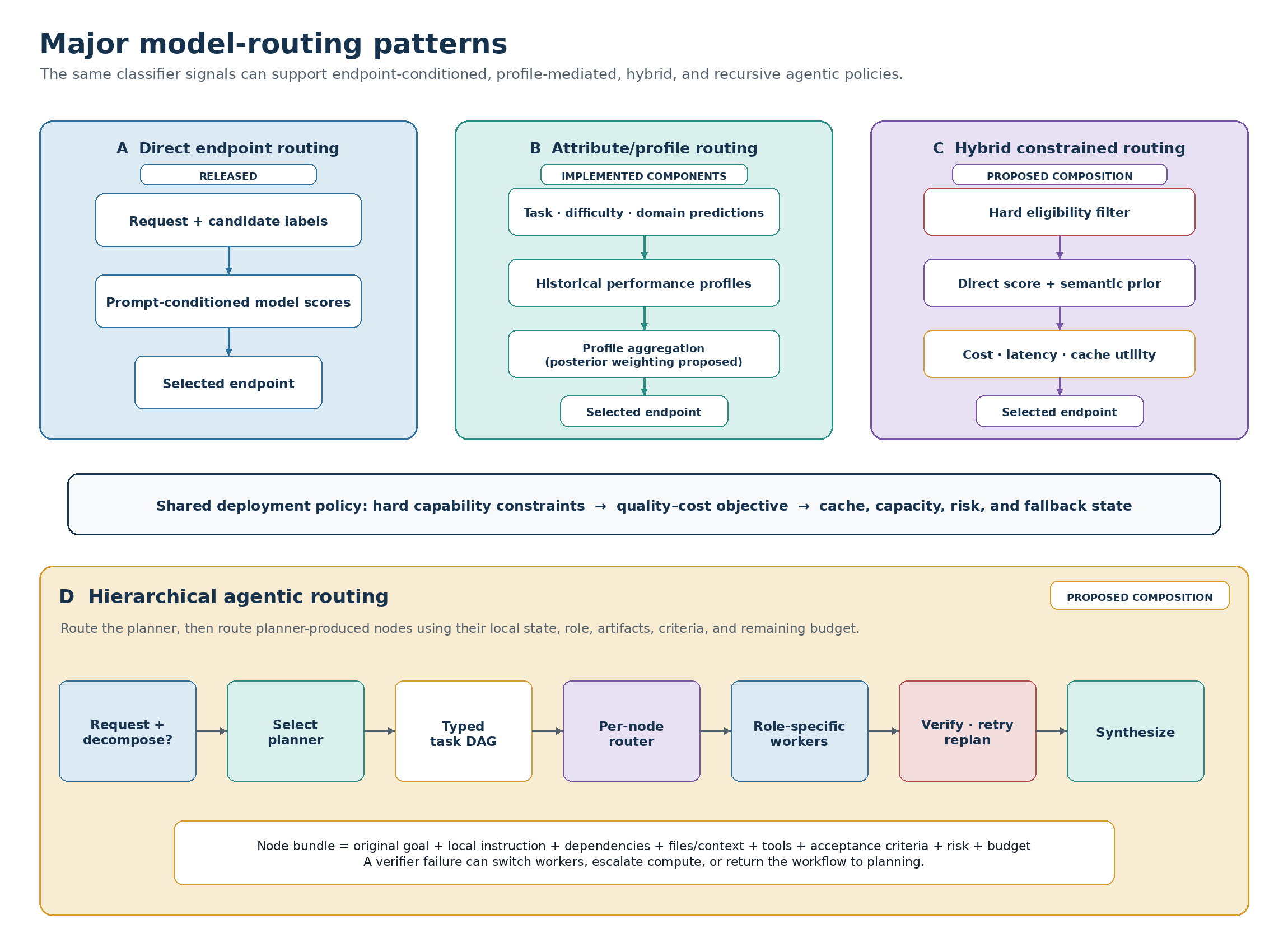}
\caption{Major model-routing patterns. (A) Direct routing scores endpoint labels from the request. (B) Attribute-mediated routing predicts task, difficulty, and domain, then consults performance profiles. (C) Hybrid routing combines direct and profile evidence before deployment policy. (D) Hierarchical routing selects a planner and routes each typed task node with verification feedback. Only the direct path has released end-to-end evidence; hard top-1 profile aggregation is implemented without end-to-end results, while posterior, hybrid, and agentic variants are proposed. Unequal endpoint coverage does not establish their relative merit.}
\label{fig:routing_patterns}
\end{figure}

The classifier supports four routing patterns, chosen according to endpoint stability, outcome-data coverage, and whether execution is a single call or a workflow. All patterns first restrict candidates to an eligible set $\mathcal E_t\subseteq\mathcal M_t$; context, modality, tools, privacy, residency, and safety remain hard constraints. Figure~\ref{fig:routing_patterns} summarizes the designs.

\subsection{Direct Endpoint Routing}

The direct pattern presents endpoint descriptions as classifier labels and produces one suitability score for each eligible endpoint:
\begin{equation}
 Q_{\mathrm{dir}}(m\mid x_t,h_{<t})
 =\sigma\!\left(s_m(x_t,h_{<t};\mathcal M_t)\right),
 \qquad m\in\mathcal E_t,
 \label{eq:direct_route}
\end{equation}
where $s_m$ is the decoder-KV classifier logit. A quality-only policy takes the highest score, while production passes the scores to Section~\ref{sec:policy}. Dynamic labels permit endpoint changes without modifying the head, but unseen endpoint names remain uncalibrated without measured outcomes. The released eight-endpoint checkpoint implements and evaluates this path.

\subsection{Attribute-Mediated Performance Routing}
\label{sec:attribute_routing}

When endpoints change faster than router training, the classifier can predict stable attributes and use a replaceable performance table. Let $\mathcal D=\{\text{task},\text{difficulty},\text{domain}\}$, let $\mathcal Z_d$ denote the categories of dimension $d$, and let $\mu_{m,d,z}$ be model $m$'s historical mean outcome for category $z$. The current optimizer selects the top category in each dimension,
\begin{equation}
 \widehat z_d=\arg\max_{z\in\mathcal Z_d}\pi_d(z\mid x_t,h_{<t}),
\end{equation}
and computes
\begin{equation}
 Q_{\mathrm{prof}}^{\mathrm{hard}}(m\mid x_t,h_{<t})
 =\frac{1}{W}\sum_{d\in\mathcal D}w_d\mu_{m,d,\widehat z_d},
 \qquad W=\sum_{d\in\mathcal D}w_d.
 \label{eq:hard_profile}
\end{equation}
Missing cells fall back to the model's overall mean. This decouples semantic prediction from the model--task table, but hard boundaries and sparse cells can mislead. The path is implemented but lacks end-to-end results.

A proposed probabilistic extension instead marginalizes attribute uncertainty:
\begin{equation}
 Q_{\mathrm{prof}}^{\mathrm{post}}(m\mid x_t,h_{<t})
 =\frac{1}{W}\sum_{d\in\mathcal D}w_d
   \sum_{z\in\mathcal Z_d}\pi_d(z\mid x_t,h_{<t})\widetilde\mu_{m,d,z}.
 \label{eq:posterior_profile}
\end{equation}
Here $\widetilde\mu$ should shrink sparse cells toward a model or task prior. Unequal endpoint coverage additionally requires confidence intervals, common-task comparisons, or missing-outcome modeling. Equation~\ref{eq:posterior_profile} is therefore proposed rather than evaluated.

\subsection{Hybrid and Adaptive Routing}

Direct scores capture request--endpoint interactions, while profiles incorporate new outcome data without retraining. After calibration, a hybrid combines both:
\begin{equation}
 Q_{\mathrm{hyb}}(m\mid x_t,h_{<t})=
 \gamma_t\,\operatorname{norm}\!\left(Q_{\mathrm{dir}}(m)\right)
 +(1-\gamma_t)\,\operatorname{norm}\!\left(Q_{\mathrm{prof}}^{\mathrm{post}}(m)\right),
 \label{eq:hybrid_route}
\end{equation}
The held-out parameter $\gamma_t\in[0,1]$ may depend on score margin, entropy, profile support, drift, or endpoint novelty, allowing reweighting or abstention. Fusion follows eligibility filtering and precedes cost, latency, cache, and capacity policy.

A confidence-gated cascade invokes an economical model and escalates when a verifier predicts failure, following FrugalGPT and AutoMix~\citep{chen2023frugalgpt,aggarwal2023automix}. A parallel portfolio instead executes top-$k$ candidates and judges or synthesizes their outputs. Both require extra resources and a deployable verifier; the $\text{router@}k$ results in Section~\ref{sec:evaluation} are oracle diagnostics, not portfolio evidence. Contextual-bandit adaptation is another proposed option, requiring controlled exploration, delayed-feedback handling, and safety boundaries.

\subsection{Hierarchical Agentic Routing}
\label{sec:agentic_routing}

An agentic system adds decisions beyond the initial prompt. A first gate chooses direct execution or decomposition; for the latter, the router selects a planner that creates a typed task graph. Each node context $c_v$ contains the objective, local instruction, predecessor outputs, files and tools, acceptance criteria, risk, and remaining budget. A role-aware policy chooses
\begin{equation}
 m_v^*=\arg\max_{m\in\mathcal E_v}U(m\mid c_v,r_v),
 \qquad
 r_v\in\{\text{plan},\text{execute},\text{verify},\text{synthesize}\}.
 \label{eq:agentic_route}
\end{equation}
Per-node routing permits different specialists for planning, execution, verification, and synthesis. Verification may accept, retry, switch models, escalate compute, or replan, subject to explicit budgets and loop limits. The 165,000 synthetic tasks in Section~\ref{sec:data} provide context, artifacts, and criteria for future evaluation, but hierarchical routing has not been evaluated end to end.

Table~\ref{tab:routing_patterns} summarizes the evidence available for each routing pattern.

\begin{table}[!htbp]
\centering
\footnotesize
\setlength{\tabcolsep}{4pt}
\renewcommand{\arraystretch}{1.13}
\caption{Routing patterns and their evidence status in this work. ``Proposed'' denotes a design, not an empirical result.}
\label{tab:routing_patterns}
\begin{tabularx}{\textwidth}{>{\raggedright\arraybackslash}p{0.18\textwidth}>{\raggedright\arraybackslash}X>{\raggedright\arraybackslash}p{0.20\textwidth}>{\raggedright\arraybackslash}p{0.23\textwidth}}
\toprule
\textbf{Pattern} & \textbf{Decision signal} & \textbf{Execution} & \textbf{Evidence status} \\
\midrule
Direct endpoint & Prompt-conditioned endpoint-label scores & One selected endpoint & Released and evaluated for eight endpoints \\
Hard attribute profile & Top-1 task, difficulty, and domain; historical table & One selected endpoint & Components implemented; no end-to-end result \\
Posterior profile & Attribute posteriors and support-aware table & One selected endpoint & Proposed extension \\
Hybrid constrained & Calibrated direct and profile scores plus policy state & One selected endpoint & Proposed composition \\
Adaptive cascade & Initial route plus confidence or verifier feedback & Sequential escalation & Proposed composition \\
Parallel portfolio & Top-$k$ candidates plus a deployable judge & Parallel calls and selection & Proposed; evaluated $\text{router@}k$ is oracle-based \\
Planner--worker & Decomposition, role, node state, and verifier feedback & Routed task graph & Proposed; agentic data is an evaluation substrate \\
\bottomrule
\end{tabularx}
\end{table}

% ========================
%        SECTION 9
% ========================
\section{Cache-Aware Routing Policy}
\label{sec:policy}

Each single-endpoint pattern in Section~\ref{sec:routing_patterns} supplies a performance signal, not an immutable product policy. After applying hard eligibility constraints, a deployment can combine that signal with incremental monetary cost and cache state. The same policy can also score planner and worker choices in Equation~\ref{eq:agentic_route}.

Let $N_c$ be cached conversation tokens, $N_n$ new input tokens, and $\widehat N_o$ expected output tokens. Each model $m$ has full input price $p_{\mathrm{in}}(m)$, cached-input price $p_{\mathrm{cache}}(m)$, and output price $p_{\mathrm{out}}(m)$ per million tokens. If $m_c$ is the current generation model and $q$ is the probability that its cache is reusable, the effective history rate is
\begin{equation}
 r_h(m)=
 \begin{cases}
 q\,p_{\mathrm{cache}}(m)+(1-q)p_{\mathrm{in}}(m), & m=m_c,\\
 p_{\mathrm{in}}(m), & m\neq m_c.
 \end{cases}
\end{equation}
The next-request cost is
\begin{equation}
 C_t(m)=\frac{r_h(m)N_c+p_{\mathrm{in}}(m)N_n+p_{\mathrm{out}}(m)\widehat N_o}{10^6}.
 \label{eq:cost}
\end{equation}
Previously billed tokens are sunk; Equation~\ref{eq:cost} prices only the next action. Switching models replays history at the new candidate's full input rate, making long cache-warm conversations naturally sticky.

The selector normalizes predicted performance and the natural logarithm of cost to $[0,1]$, inverts cost so larger is better, and computes
\begin{equation}
 U_t(m)=\alpha U_{\mathrm{perf}}(m)+(1-\alpha)U_{\mathrm{cost}}(m).
 \label{eq:utility}
\end{equation}
Normalization uses historical mean $\pm3\sigma$ when available, configured bounds otherwise, and candidate min--max as a fallback. Degenerate ranges yield neutral utilities. Ties break by higher raw performance, lower cost, and then model identifier, making the decision deterministic. One configurable multi-signal example assigns weights 0.60 to suitability, 0.15 to task fit, 0.15 to difficulty fit, and 0.10 to reasoning fit. These weights are illustrative, not learned constants or universal defaults.

Safety, privacy, data residency, tool permissions, and context feasibility should remain outside Equation~\ref{eq:utility} as hard filters. A high utility must never compensate for ineligibility.

% ========================
%        SECTION 9
% ========================
\section{Evaluation}
\label{sec:evaluation}

\subsection{Metrics}

We evaluate the released model at two levels: per-family classification and end-to-end fixed-versus-router comparisons. The former measures the prediction heads; the latter measures realized downstream task quality.

For multi-label model routing, we compute per-candidate precision, recall, and F1 at threshold 0.5 and take their unweighted macro average. For single-label auxiliary families, macro F1 is averaged across classes. A separate 3,192-record evaluation uses a different aggregation and obtains model-routing F1 0.805, whereas the explicit per-candidate macro average is 0.759. Because the protocols differ, we keep these values separate and do not interpret the difference as a temporal regression.

\begin{figure}[!htbp]
\centering
\includegraphics[width=0.94\textwidth]{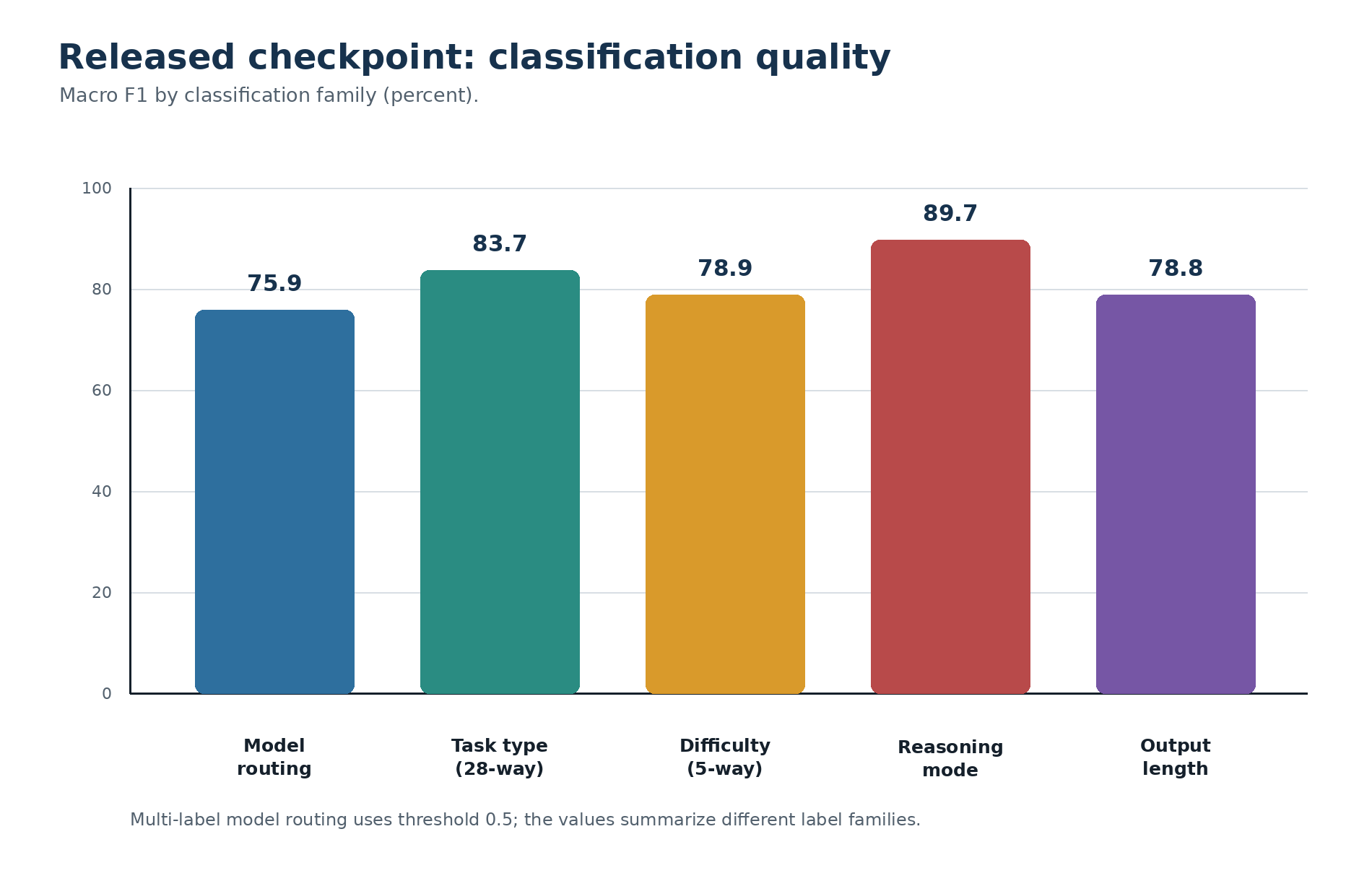}
\caption{Macro F1 by released-checkpoint classification family. The values summarize distinct label families and are not averaged into a single system score.}
\label{fig:head_metrics}
\end{figure}

Figure~\ref{fig:head_metrics} and Table~\ref{tab:head_metrics} summarize the released-checkpoint classification families.

\begin{table}[!htbp]
\centering
\small
\caption{Released-checkpoint metrics. The hallucination score is a directional response signal.}
\label{tab:head_metrics}
\begin{tabular}{lrrl}
\toprule
\textbf{Family} & \textbf{Classes} & \textbf{Macro F1} & \textbf{Decision view} \\
\midrule
Model suitability & 8 & 0.759 & multi-label, threshold 0.5 \\
Task type & 28 & 0.837 & single-label \\
Difficulty & 5 & 0.789 & single-label, ordinal \\
Reasoning mode & 2 & 0.897 & single-label \\
Expected output length & 7 & 0.788 & single-label, ordinal buckets \\
Hallucination & 2 & $\sim$0.65 & directional response signal \\
\bottomrule
\end{tabular}
\end{table}

\FloatBarrier
\subsection{Model-Routing Classification}

Table~\ref{tab:model_routing} shows the released eight-candidate model-routing result. Precision is comparatively stable (0.739--0.840), whereas recall ranges from 0.650 to 0.950. This matters operationally: a high-recall generalist appears in more positive sets, while a specialist can have high precision but be missed on tasks it could solve. Threshold tuning should therefore be candidate- and objective-aware rather than assumed universal.

\begin{table}[!htbp]
\centering
\small
\caption{Per-candidate routing precision, recall, and F1. Endpoint names are shortened only for display.}
\label{tab:model_routing}
\begin{tabular}{lrrr}
\toprule
\textbf{Candidate} & \textbf{Precision} & \textbf{Recall} & \textbf{F1} \\
\midrule
Gemma 4 31B & 0.8064 & 0.9501 & \best{0.8723} \\
Llama 4 Maverick & 0.7892 & 0.7529 & 0.7706 \\
GPT-OSS 120B & 0.8396 & 0.7063 & 0.7672 \\
Coder & 0.8140 & 0.7000 & 0.7527 \\
Llama 3.3 70B & 0.7638 & 0.7239 & 0.7433 \\
DeepSeek V3.1 & 0.7475 & 0.7220 & 0.7346 \\
MAGPiE & 0.7394 & 0.7234 & 0.7313 \\
Qwen3 32B & 0.7505 & 0.6499 & 0.6966 \\
\midrule
Macro mean & 0.7688 & 0.7411 & 0.7586 \\
\bottomrule
\end{tabular}
\end{table}

The 28-way task taxonomy is near-perfect on semantically distinct classes such as analysis, code, generation, information extraction, QA, reasoning, summarization, and translation. The principal failure is problem solving (F1 0.127), followed by math-and-reasoning (0.537), decision support (0.599), information retrieval (0.626), and evaluation (0.642). These failures motivate the ontology's explicit boundary notes and its decision to move interaction properties out of the intent hierarchy.

Difficulty is ordinal. The released-checkpoint evaluation obtains macro F1 0.789, with most errors between adjacent hard and extra-hard levels. The separate 3,192-record evaluation gives top-1, top-2, and top-3 hit rates of 0.507, 0.748, and 0.873, respectively. Because the protocols differ, we do not merge the values; both motivate distance-aware ordinal error alongside exact class F1.

\FloatBarrier
\subsection{End-to-End Routing}

\begin{figure}[!htbp]
\centering
\includegraphics[width=0.94\textwidth]{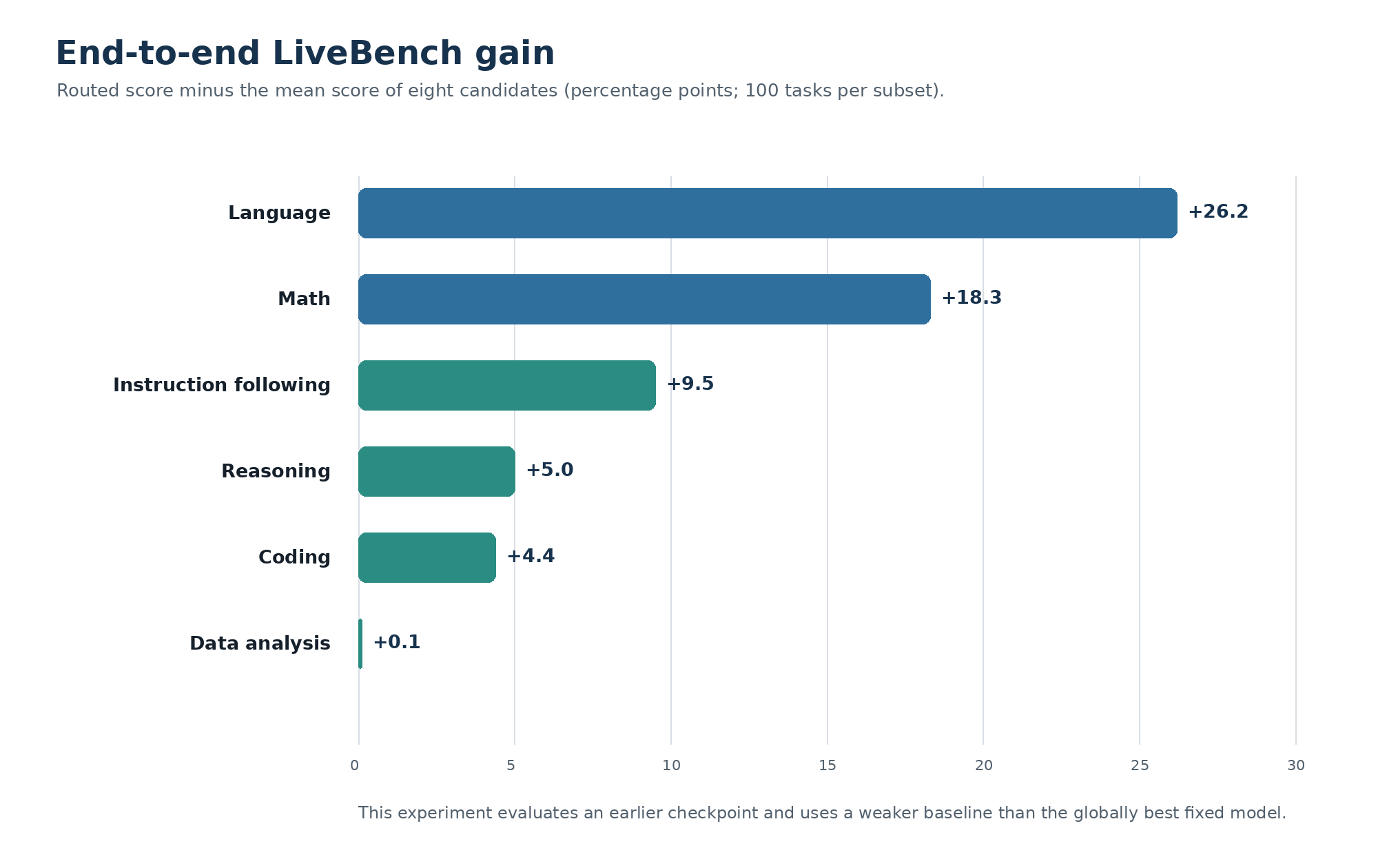}
\caption{LiveBench gain over the mean of eight candidate scores. This experiment evaluates an earlier system checkpoint and is not a final-checkpoint claim.}
\label{fig:livebench}
\end{figure}

We use two end-to-end baselines. The first compares the routed top-1 score with the arithmetic mean of eight candidate scores on six LiveBench subsets~\citep{white2024livebench}. This measures improvement over a uniformly sampled candidate, not over a strong fixed policy. Table~\ref{tab:livebench_mean} and Figure~\ref{fig:livebench} give the results.

\FloatBarrier

\begin{table}[!htbp]
\centering
\small
\caption{Prior LiveBench routed score versus the mean candidate score, 100 tasks per subset.}
\label{tab:livebench_mean}
\begin{tabular}{lrrr}
\toprule
\textbf{Subset} & \textbf{Router} & \textbf{Mean of 8} & \textbf{Gain} \\
\midrule
Language & 0.779 & 0.517 & +0.262 \\
Math & 0.738 & 0.555 & +0.183 \\
Instruction following & 0.863 & 0.768 & +0.095 \\
Reasoning & 0.601 & 0.551 & +0.050 \\
Coding & 0.500 & 0.456 & +0.044 \\
Data analysis & 0.540 & 0.539 & +0.001 \\
\bottomrule
\end{tabular}
\end{table}

The second baseline is stricter. \texttt{fixed@k} selects the globally top-$k$ candidates by mean score on the same evaluation records; \texttt{router@k} forms a per-task top-$k$ shortlist and takes the best \emph{realized} outcome within that shortlist. At $k=1$, both are deployable single-model policies. For $k>1$, \texttt{router@k} is an oracle-within-shortlist diagnostic unless an online selector is specified. The globally fixed ranking is selected in-sample and can therefore be optimistic.

The full evaluation contains 1,500 tasks across 13 benchmarks. The dataset-level analysis uses a selected 1,000-task subset comprising LiveBench and non-LiveBench datasets with a positive router gain at one or more depths. On this subset, fixed/router scores are 0.696/0.707 at $k=1$, 0.788/0.794 at $k=2$, and 0.837/0.824 at $k=3$, giving gains of $+0.012$, $+0.007$, and $-0.013$. Table~\ref{tab:strict_top1} gives the top-1 dataset values.

\begin{table}[!htbp]
\centering
\small
\caption{Top-1 fixed-versus-router diagnostic. ``All shown'' is sample-weighted over the selected 1,000-task subset, not the full 1,500-task evaluation.}
\label{tab:strict_top1}
\begin{tabular}{lrrr}
\toprule
\textbf{Dataset} & \textbf{Fixed@1} & \textbf{Router@1} & \textbf{Gain} \\
\midrule
BBEH~\citep{kazemi2025bbeh} & 0.830 & 0.830 & +0.000 \\
MuSR~\citep{sprague2023musr} & 0.677 & 0.683 & +0.007 \\
LiveBench coding & 0.470 & 0.500 & +0.030 \\
LiveBench data analysis & 0.560 & 0.550 & $-0.010$ \\
LiveBench instruction following~\citep{zhou2023ifeval} & 0.834 & 0.854 & +0.020 \\
LiveBench language & 0.659 & 0.726 & +0.067 \\
LiveBench math & 0.738 & 0.738 & +0.000 \\
LiveBench reasoning & 0.834 & 0.824 & $-0.010$ \\
\midrule
All shown & 0.696 & 0.707 & +0.012 \\
\bottomrule
\end{tabular}
\end{table}

These results support a narrower conclusion than ``routing always wins.'' The router is most useful where candidate outcomes disagree and request semantics reveal that disagreement; it adds little where candidates cluster and can underperform a fixed baseline. Because the 1,000-task analysis subset omits five benchmarks and lacks uncertainty intervals, repeated seeds, endpoint versions, and held-out fixed-policy selection, it does not support a global significance claim. We do not report downstream comparisons for the unequally covered eleven-endpoint collection.

% ========================
%        SECTION 11
% ========================
\section{Discussion}
\label{sec:discussion}

\router{} demonstrates a practical benefit of treating model selection as compact, dynamic-label classification rather than as another generation task. A small router can score a changing endpoint roster, reuse conversational context through the decoder-KV path, and expose task, difficulty, reasoning, and output-length signals to an explicit deployment policy. This separation makes eligibility, safety, price, latency, and cache reuse controllable rather than implicit in a single model prediction. The task ontology and 165,000 synthetic tasks further extend this interface from conventional prompts to tool use, repository work, and agentic workflows.

The empirical advantage is nevertheless conditional. Routing helps most when candidate models disagree and the request contains signals that predict those differences; a strong fixed model remains competitive when outcomes are similar. Current end-to-end evidence covers direct routing on a selected subset, while unequal coverage prevents a balanced comparison across all eleven expanded endpoints. Synthetic generation, verifier design, and \texttt{gpt5.6-sol} judging can introduce author-model, evaluation-shortcut, and judge biases~\citep{zheng2023judging,verga2024replacing}. The full task corpus and outcome matrix are also not yet public. The router should therefore be treated as a performance signal inside a constrained policy, with safety, privacy, residency, and tool authorization enforced independently.

Future work should add more model families, sizes, modalities, context lengths, tool-use capabilities, and price--latency tiers, with every endpoint evaluated on shared, versioned task strata. The four routing patterns should then be compared on the same outcome matrix: direct endpoint prediction, attribute-mediated performance profiles, hybrid scoring, and hierarchical planner--worker routing for agentic tasks. Evaluation should measure realized quality, cost, latency, calibration, endpoint-drift robustness, and regret to an oracle. Logged propensities, controlled exploration, and shadow evaluation are also needed to improve the router without reinforcing its own selection bias. This program would reveal where each routing pattern is useful and turn the ontology expansion into a reproducible benchmark.

% ========================
%        SECTION 12
% ========================
\section{Conclusion}
\label{sec:conclusion}

\router{} frames model selection as lightweight, dynamic-label classification over an evolving endpoint roster. The approximately 0.6B-parameter checkpoint combines request-conditioned model scoring with task, difficulty, reasoning, and output-length predictions, while its decoder-KV path supports repeated routing over conversational state. On the selected 1,000-task evaluation subset, direct routing reaches a top-1 score of 0.707 compared with 0.696 for the strongest fixed endpoint. Together with the per-head classification results, this shows that a compact router can exploit predictable model specialization without requiring an additional generative decision step. The mixed per-benchmark gains also clarify that this advantage depends on meaningful disagreement among candidate models.

The task ontology, 150,000 verifier-scored tasks, and 15,000 \texttt{gpt5.6-sol}-judged tasks broaden the routing target toward realistic applications and agentic workflows. The routing-pattern framework connects this data to direct, attribute-mediated, hybrid, and hierarchical planner--worker strategies while keeping deployment constraints separate from learned predictions. This work therefore provides a foundation rather than a final universal leaderboard. A paired and versioned outcome matrix across more models, followed by realized quality--cost--latency evaluation of each routing pattern, is the central requirement for establishing when routing delivers reliable operational value.

\section*{Acknowledgements}

This work was developed jointly by SCX.ai Holdings Limited and Knowledgator.

\bibliography{references}

\appendix

% ========================
%        APPENDIX A
% ========================
\section{Ontology Families}
\label{app:ontology}

Table~\ref{tab:ontology_families} lists the complete top-level ontology. Its deliberately regular construction assigns five task types, 15 subtypes, and 51 synthetic example specifications to each family.

\small
\begin{longtable}{@{}p{0.07\textwidth}p{0.52\textwidth}rrr@{}}
\caption{Complete task-family inventory.}\label{tab:ontology_families}\\
\toprule
\textbf{ID} & \textbf{Family} & \textbf{Types} & \textbf{Subtypes} & \textbf{Tasks} \\
\midrule
\endfirsthead
\toprule
\textbf{ID} & \textbf{Family} & \textbf{Types} & \textbf{Subtypes} & \textbf{Tasks} \\
\midrule
\endhead
F01 & Information Retrieval and Discovery & 5 & 15 & 51 \\
F02 & Question Answering and Explanation & 5 & 15 & 51 \\
F03 & Summarization and Synthesis & 5 & 15 & 51 \\
F04 & Information Extraction and Structuring & 5 & 15 & 51 \\
F05 & Classification and Tagging & 5 & 15 & 51 \\
F06 & Clustering and Organization & 5 & 15 & 51 \\
F07 & Analysis and Insight & 5 & 15 & 51 \\
F08 & Comparison and Benchmarking & 5 & 15 & 51 \\
F09 & Evaluation and Scoring & 5 & 15 & 51 \\
F10 & Critique and Review & 5 & 15 & 51 \\
F11 & Fact-Checking and Verification & 5 & 15 & 51 \\
F12 & Logical, Mathematical, and Causal Reasoning & 5 & 15 & 51 \\
F13 & Problem Solving and Troubleshooting & 5 & 15 & 51 \\
F14 & Forecasting and Estimation & 5 & 15 & 51 \\
F15 & Decision Support and Recommendation & 5 & 15 & 51 \\
F16 & Planning, Scheduling, and Optimization & 5 & 15 & 51 \\
F17 & Writing, Rewriting, and Translation & 5 & 15 & 51 \\
F18 & Content Generation and Visualization & 5 & 15 & 51 \\
F19 & Programming and Software Engineering & 5 & 15 & 51 \\
F20 & Data, Quantitative, and Scientific Computing & 5 & 15 & 51 \\
F21 & Agentic Tool Use and Workflow Execution & 5 & 15 & 51 \\
F22 & Conversation and Customer Support & 5 & 15 & 51 \\
F23 & Safety, Risk, and Compliance & 5 & 15 & 51 \\
\midrule
\textbf{Total} & & \textbf{115} & \textbf{345} & \textbf{1,173} \\
\bottomrule
\end{longtable}
\normalsize

The 30 domains are grouped into Public, Social, and Education Services (3); Health and Life Sciences (2); Financial, Legal, and People Services (5); Digital Technology and Communications (6); Industry, Infrastructure, and Property (5); Commerce, Logistics, and Hospitality (4); Environment and Primary Production (2); and Research, Media, and Sport (3).

\end{document}